\documentclass[runningheads]{llncs}
\usepackage[utf8]{inputenc}
\usepackage{graphicx}
\begin{document}

\title{An Overview of Blockchain Integration with Robotics and Artificial Intelligence\thanks{This work was partially supported by the Tezos Fundation through a grant for project Robotchain.}}
\titlerunning{An Overview of Blockchain Approaches}
%
%
\author{Vasco Lopes \and
Luís A. Alexandre}
\authorrunning{V. Lopes and L. A. Alexandre}
%
\institute{Departamento de Informática, Universidade da Beira Interior\\and Instituto de Telecomunicações\\
Rua Marquês d’ Avila e Bolama, 6201-001, Covilhã, Portugal\\
\email{vascoferrinholopes\{at\}gmail.com\\ luis.alexandre\{at\}ubi.pt}}
%
\maketitle              
\begin{abstract}
Blockchain technology is growing everyday at a fast-passed rhythm and it's possible to integrate it with many systems, namely Robotics with AI services. However, this is still a recent field and there isn't yet a clear understanding of what it could potentially become. In this paper, we conduct an overview of many different methods and platforms that try to leverage the power of blockchain into robotic systems, to improve AI services or to solve problems that are present in the major blockchains, which can lead to the ability of creating robotic systems with increased capabilities and security. We present an overview, discuss the methods and conclude the paper with our view on the future of the integration of these technologies.

\end{abstract}
\section{Introduction}



Blockchain technology was first introduced by Satoshi Nakamoto \cite{Nakamoto2008} alongside with the crypto-currency Bitcoin. Both have grown in terms of adoption, value and usage \cite{BlockchainCharts,Blockchaingrowth}, but the value of blockchain is not only to hold crypto-currencies but to allow the integration of a huge panoply of systems over the same platform in a decentralised and secure way. Ethereum \cite{wood2014ethereum}, proposed in 2013, introduced new features to the blockchain technology, such as smart-contracts, that changed the whole game for this technology, allowing it to integrate more services and have more value to many industries and academic fields. Although these projects opened many doors, they are still lacking on some essential characteristics, like the energy consumption and the speed that a block takes to be validated. With these problems in mind, every year thousands of new ideas and services to work with the blockchain technology are proposed, but there isn't a unique solution for all possible applications and that encourages the development of new work. This lack of a single solution is well seen in the robotics field, where integration with blockchain is still low and there aren't many approaches that show how both technologies can overcome challenges together. The introduction of blockchain in robotic systems could solve many problems that those systems face. The first problem that it can solve is security, as many of the systems have problems of trust between robots and data integrity, blockchain can provide a reliable peer-to-peer communication with security measures over a trustless network. Another advantage of this integration is the possibility to make distributed decisions since the blockchain can ensure that all participants of a decentralised network share identical views of the world. This assurance can allow the system to reach an agreement over the whole network and to have global collaboration between the robots.

In this paper, we give an overview on the current state-of-the-art related to blockchain technology integrated with Robotics or Artificial Intelligence (AI) and how the market is shifting, by studying the major systems that implement services with Robotics, and/or AI. We value the AI aspect of the works because robotics is heavily based on AI systems, and if a system can work towards improving those systems, the more likely is that robotics can be integrated there and benefit from that platform.

The remainder of this paper is organised as follows: Section \ref{sec:overview} explains, reviews and compares the different methods, according to what their fundamental base is. Section \ref{sec:disc} presents our ideas and discusses the reviewed methods. Section \ref{sec:conc} summarises the main ideas presented throughout the paper.

\section{Overview}
\label{sec:overview}
In this section we present work that has been done that tries to integrate blockchain into robotics, AI or that adds significant features that we believe to be crucial for robotic systems, like enhanced security, the ability to change the blockchain protocols without the need for a hard-fork or the ability to validate more transactions per second than most used blockchains, since this is a major concern to register robotic events. The section is split into 3 subsections, the first one, Robotics, explains some work that has been done entirely on improving robotic systems with blockchain or that have robotic services that are already implemented. The second subsection is Artificial Intelligence, and this is the biggest subsection of the three since it shows the work that has been done with blockchain and AI that has some implementation of robotics or that can be integrated with robotic systems. The third and last subsection presents work that has neither AI or robotics as the main focus but that can still give robotic systems a leverage.

\subsection{Robotics}

The work conducted by Bruno Degardin and Luís A. Alexandre (as supervisor) \cite{BrunoDegardinLA}, shows how to create a blockchain and use it to store robotic events. This idea allows the creation of smart-contracts that use information acquired on the wild by different robots (possibly from different manufacturers) and have action-triggers based on the contracts that are stored and verified on the blockchain. This can ultimately improve productivity in a factory and reduce the time spent on doing tasks like refilling the screws for a robot that used the blockchain to indicate that he needs more screws to continue is work.

As a follow-up to this work, and now using Tezos' technology, M. Fernandes and L. Alexandre \cite{RobotChain} propose the creation of a blockchain for robotic event registration that takes advantage of the improved security provided by the formal verification embedded into Tezos (more details on Tezos' technology, in section \ref{subsec:others}). This on-going work will support smart contracts to run AI code over the blockchain, where these smart-contracts are proved correct (to do exactly what their specification defines). They also plan to adapt the blockchain to support many more events per second than the current specification enables, to allow for the system to deal with a large number of interacting robots.

Aitheon \cite{Aitheon} is a platform that has been developed for over a decade and has recently adopted a blockchain technology based on the ERC-20 standard. Their goal is to have a complete platform that can reduce the number of time-consuming tasks that developers have to endure, like organising documents. Their solution is a platform with both AI and Robotics that can provide automation for specified business processes. The platform they built has 5 modules. The first, an AI module, tries to retrieve information about frequently performed tasks and takes actions to automate those tasks. The second module, called Digibots, is very similar to the first one, but this module focus on automating programming tasks, like back-end solutions and data-driven problems. The third module, Mechbots, is focused on helping businesses integrate robotic automation to increase efficiency and productivity. The last two modules, Aitheon Specialists and Pilots are human specialists that conclude the tasks that can't be fully automated and also provide robot supervision. In short, Aitheon provides a PaaS that provides a platform easy to be used by a wider audience (even with no-tech skills), that aims to automate time-consuming tasks, that is capable of integrating new robots without effort and at the same time, reduce the payment latency and increase security by using an ERC-20 token that can be used by an AI algorithm to do payments.

Work conducted by Eduardo Castelló Ferrer \cite{Ferrer2016} presents the benefits of combining the blockchain technology with robotics, especially swarm robotics and robotic hardware. The advantages of robotic swarms are how easy it is to scale and the robustness to failure. These advantages come from the fact that members of these swarms are distributed. The author shows that the research on the swarm robotics field has grown from approximately 10 papers published in 2000 to more than 250 in 2015. In the industrial sector we can also see how this market is growing and allowing companies to achieve higher productivity, which is the case of AmazonRobotics \cite{AmazonVision} that has been on the news showcasing its army of robots that cooperatively work to manage their warehouses \cite{AmazonNEWS}. Most of the robotic swarms only use local information, this means that a robot only has information about itself and/or robots that are close to it, but the integration of blockchain in these systems can give the robots global information, which can be useful for different applications. The blockchain can also improve the speed of how the system changes the behaviour, since having global information allows the whole system to quickly change behaviour to address specific robot needs. This can also be done by a controller robot that evaluates the system state by using blockchain information and commits to it the changes to be made. In short, the integration of blockchain into swarm robotic systems may lead to huge progress in the way that robots can change their state and behaviour depending on global information they can trust because of the blockchain security measures, leading to higher productivity and easier maintenance.

\subsection{Artificial Intelligence}

Recently, SingularityNET \cite{SingularityNET} has overwhelmed the media with monumental ideas and objectives. Their premise is to build a decentralised platform that allows anyone to create, share, and monetise AI services at scale. They intend to democratise AI by creating a marketplace, over a blockchain technology based on ERC20 standard \cite{ERC20} over Ethereum \cite{wood2014ethereum}. This allows the authors to create and use a specific token for their platform, called AGI (a reference to, but not the same as Artificial General Intelligence (AGI)), that can be bought and traded with Ether (Ethereum token) or other ERC20 tokens. More concisely, the idea behind SingularityNET is both the creation of a platform where developers can monetise the algorithms they export to the SingularityNET's blockchain and also the integration of a high-level API that should allow everyone to easily build and deploy systems with numerous algorithms that are allocated in the blockchain. The API system works on the basis that a user is willing to pay to have an algorithm do a specific job. The algorithm chosen to conduct the routine may need other sub-routines, which can be done by other algorithms in the blockchain. These sub-routines are paid by the API that is handling the principal routine, which deducts the earnings of the first algorithm but allows it to perform an extensive work, even though it may not know how to do it. For the sake of exemplification, imagine that you want to create a software capable of detect profanity words in the news, you use the SingularityNET API and select one of the suggested algorithms for that job, but most certainly, some news will be in a foreign language and need translation. This is where the API handles the subcontracting of other algorithms to handle that sub-problem. The goal of the developers is that this will allow every user to easily use AI, despite the fact that the algorithms might be a black-box for the user. SingularityNET wants to provide a mechanism to create a full network of different algorithms where which one of them is only built to solve one specific problem. This has as a foundation previous work conducted by some of the same authors \cite{Hart:2008:OSF:1566174.1566223}. This foundation allied with the decentralised network of algorithms allows the creation of some sort of AGI, where a unique system can perform multiple and complex tasks. The authors like to reference the expressive robot Sophia \cite{Sophia} as one of the cases that serve both as a test and as a client of this AGI idea.

In the White paper by Namahe \cite{Namahe}, a real-world application of how blockchain and AI can work together towards a more transparent chain in the fashion world is shown. Their goal is to create a platform, based on the blockchain technology to aggregate all the different tiers and stakeholders of this industry into one single platform. This idea allows them to reduce costs to companies since they have everything they need in a single platform, ranging from sellers to buyers, and to enforce the transparency of all trades, ensuring that the workers are paid at least the minimum wage and also contribute to reducing child labour. The AI coupled to this blockchain has the job of acting as an invisible tracking machine that proactively identifies issues before they happen and suggests mitigation's to keep the supply chain running efficiently.

Numerai \cite{Numerai} has built a unique concept over blockchain. Their concept allows data scientists to make predictions on one of the toughest and more competitive markets, the stock market. They built a system that tries to nullify the fact that the stock market has very slow progress in the fields of ML. This slow progress happens because the data is very sensitive and only a few data scientists have access to the raw data. Numerai overcomes this challenge by an abstraction of the data. This abstraction is built with a unique algorithm of ciphering the stock market data without losing the prediction aspect inherent to the data. With this new way of presenting the data, they are able to share it with their community and allow everyone to participate with his or her prediction algorithms onto that data. This, in the developer's words, is a way of overcoming the human bias and overfitting of the data. With the ensemble of prediction algorithms, Numerai holds a hedge fund that uses the prediction algorithms (tested on new data that developers don't have access to, to reduce the bias) to invest the capital, depending on the results and the direction that the algorithms take them. The blockchain is present in this project so that they can have a decentralised system of predicting algorithms and to have a specific ERC-20 token, called Numeraire. This token is used by the users to participate in the weekly competitions, where they bet Numeraires in their algorithm's performance. This serves the purpose of defining the "confidence" that the data scientist has on his algorithm, and a higher confidence can lead to higher earnings if the algorithm passes the tests.
DeepBrain Chain \cite{DeepBrainChain} launched a platform where they aim to create a cloud of graphic cards. With the recent developments in the deep learning area, large amount of data is required to train the most proficient models, but this implies that the developers have access to multiple graphic cards. But, even if developers have access to some cards, the time taken to train a Neural Network (NN) such as a Convolution Neural Network (CNN) can take days. So, DeepBrain chain created a way of allowing the miners of cryptocurrencies to mine the blockchain they created on their platform, but, instead of doing a typical proof-of-work, they are actually allowing others to use their graphic cards. This way, researchers and developers can have access to thousands of graphic cards. The blockchain ensures that all the parties comply with their agreement and that the data that is being used to train the NNS is not accessed by people without permission. This is all done using smart-contracts over the blockchain with an ERC-20 compliance token. With this platform the miners get more profit for their smart-contract mining and graphic card renting, the users spend less money because they have no need for expensive physical equipment. This platform can easily be used to build an AI ecosystem dedicated to robotics.

Other project that has some similarities with DeepBrain Chain is Neuromation \cite{Neuromation}. The underlying idea is the same, they try to leverage the distributed computational power associated with the blockchain to allow developers to train NNs. The main difference is that the Neuromation Market Platform unites the scientific community, market resources and commercial and private entities in one single place. This market allows anyone to buy or sell datasets, models and AI services, like labelling of data. The decentralised computing power also allows this marketplace to sell AI models or train a brand-new one. Neuromation has practical applications already in place by their partners, mainly in medical detection problems and industrial robots by training models with synthesised data.

Eligma \cite{Eligma} is a company that aims to transform the way people interact with products, allowing them to easily buy, track and sell all kinds of products. They developed a platform that has two cores. The AI core, consists of a chatbot, that is designed to help users to interact with the platform, a product-matching algorithm, that depending on user inputs and preferences matches them with different products that are on the platform and finally, a value-prediction algorithm that does a present and future estimation of all products that are on sale. The second core is the blockchain, where they use the Ethereum blockchain for protection and as a side-chain for data storage, reducing the costs of maintaining the main chain with all the data. The blockchain also allows the platform to have a decentralised sales point, where two parties have a mutual agreement on some product and this becomes a smart-contract over the blockchain, introducing more security and ensuring that both parties comply with the contract. The blockchain is based on the ERC-20 standard, which allows the developers to introduce the idea of making a payment over the blockchain possible with multiple currencies. This means that a user can pay for a product with both Bitcoin, Ethereum token, ELI (Eligma token), united states dollar or make a crypto-based loan.

The work done by TraDove \cite{TraDove} is very similar to the one done by Eligma but focused on businesses. It aims to provide a way for different businesses to promote their products and match with sellers and buyers that have similar interests and profiles. This matching is done by AI technologies. With the blockchain layer, they introduce a payment that lowers the costs regarding taxes when sending money to multiple countries, but, they aim to use this layer as a way to induce transparency about the exchanges and allowing the sellers to advertise their products, with the B2BCoin (TraDove token), to specific targets.

AdHive \cite{AdHive}, created a platform focused on the advertisement market. The platform is heavily based on AI algorithms and blockchain. It has two types of users. The first type is the advertisers, that submit videos or other pieces of information that they want to be advertised on social media and blogs and specify a maximum budget for the advertisement. This request for advertisement enters the blockchain as a smart-contract that retains the budget until all the conditions are met or until the contract is cancelled. The second type of users are the influencers, that have the means to advertise the information. The AI algorithms verify if the influencers put the correct information on their dissemination media and apply the information they acquire into the blockchain so that the smart-contracts are fulfilled.

Matrix \cite{Matrix} is a project that wishes to solve four problems with the blockchain technologies: the barriers associated with smart contracts due to the fact that these are usually done in specialised programming languages, the lack of security associated with the smart contracts, the slow transaction speed and the inflexibility in managing and updating blockchains. The first problem is solved in this platform by introducing a NN that is able to automatically convert simple scripts that contain information regarding the transaction conditions and the inputs and outputs to a smart contract. This enables all users to create smart contracts. The second problem is created by having smart-contracts calling external functions and by having an open and decentralised approach to the smart-contracts. Matrix solves this problem by providing 4 extra components, a rule-based semantic and syntactic analysis engine for the smart contracts, a formal verification toolkit to provide the security properties of a smart contract, an AI-based engine that detects problems in the transaction models and checks their security and lastly, a deep learning based platform for dynamic security verification and enhancement. The third problem is a well discussed one, given the fact that Bitcoin takes about 60 minutes to confirm a transaction and Ethereum can achieve about 15 transactions per second, major sellers and some real-world problems can't leverage the blockchain technology. Matrix accelerates the transfer per second by allocating the PoW processing in a delegation network, which incurs smaller latency because the number of nodes is drastically reduced. The developers select the delegation network in a random way in the sense that the probability of a node to be selected is proportional to its PoS. The platform claims to have achieved 50 thousand transactions per second on their testnet. The fourth problem presented by the developers is overcome by giving access control and routing services to allow integration of private chains into a common public chain. This integration allows information to flow between the chains to meet the user criteria. Matrix also uses a reinforcement-learning algorithm to optimise the parameters regarding the transactions. The last thing that Matrix introduced on their platform is a new mining mechanism in which miners perform Markov Chain Monte Carlo computing, reducing both the time spent on each transaction and the waste of energy.

Marketplaces are one of the most common applications of blockchain technology. The AI Technology Network (ATN) \cite{ATN} is one of those that leverages the blockchain to propagate AI services over a marketplace. ATN is built over a smart-contract enabled blockchain so it can provide datasets and AI algorithms in a secure and trustworthy way. The developers introduced an open interface blockchain platform that enables to solve some interoperability issues between AI and smart-contracts that complies with the major standards. ATN makes it possible for users, developers and sellers to share services without concerns about security using a robust system that integrates cross-chain (multiple blockchains, specially Ethereum, Qtum, RSK and others) compatibility and a platform design that ensures the interoperability among the services on the marketplace. The ATN platform encrypts all the services that are integrated into the platform so they can ensure that the developers that integrate services on their marketplace are remunerated for their work and have the assurance that their product is secured. This platform is open-source to encourage the development of new applications.

Cortex \cite{Cortex} propose a solution for the problems associated with smart-contracts. The fundamental idea is a blockchain where users can augment their smart-contracts with the AI system proposed by Cortex. This enables the creation of decentralised applications that can react to both external and internal variables in the blockchain. Cortex built this system over a public blockchain of their own but in the early stages, they have integration with an ERC-20 token. The system also has other layers upon the blockchain. One of those layers is a platform that encourages developers, through token rewards, to upload Machine Learning (ML) models and to optimise existing ones. This collective effort may lead to achieving AGI over the Cortex platform, allowing decentralised AI applications to be very robust and highly complex. Cortex has its own virtual machine which is compatible with the Ethereum Virtual Machine (EVM), allowing it to work with Ethereum smart-contracts. Users can pay Cortex tokens to use the Cortex virtual machine in order to use AI algorithms that are under the Cortex platform. As consensus over the Cortex smart-contracts, all nodes must agree on the outcome of the inferred result. 

Application of blockchain and AI in the medical field can lead to great improvements both for patients and for doctors. Some work on this has been conducted by NAM \cite{NAM}. NAM is trying to revolutionise the state of medical care, with focus on Japan. It's a project that integrates these technologies to try to reduce medical costs, helping patients improve their health and allowing doctors to grasp the patient's information regarding a disease or a treatment. This platform, called NAM Chain, intends to build several AI services. The first one is a consulting bot that can be used by any person. It works by introducing the symptoms on the app and the bot will say how urgent the symptoms are. The second one is a set of prediction models to detect diseases by scanning medical reports. The third one is a service that serves as a nutritionist, recommending healthy food based on the user's lifestyle and constitution. The fourth one is the basis of the platform, where all the services work, which the developers call "a system of medical records with deep learning and blockchain" \cite{NAM}. This blockchain is based on the Ethereum network and blockchain but the developers want to migrate to a private blockchain in the future. The way to fight the resistance of integrating medical information, namely large images, in the blockchain is by giving the miners NAM Coins (NAM Token), which will be usable in clinics and in the AI services that the platform provides. They state that they developed a Top-K shortest path distance from sender to receiver to speed up the package transmission. The developers also assure that they work in security and that they want every report to be secure but they are very vague on explanations and don't have a clear explanation on how they will prevent people from accessing other peoples' information and how they will assure that the information on the blockchain can't be seen by those who should not have permission to see it.

In the work done by Jianwen Chen et al. \cite{Chen} a novel method to solve the problems associated with consensus on blockchain is proposed. In most of the working systems based on blockchain technology, the consensus is reached by PoW, where every node receives a crypto-puzzle that needs to be solved or PoS that takes into account the amount of stake each node has. Both of these approaches can consume large amounts of electricity, both can take a long time to reach consensus and some approaches may even lead the whole blockchain to a centralised system. The proposed framework to solve these problems is based on AI technology. The framework works as follow: first, a CNN calculates the average transaction of each node, then, statistics about the threshold values of the average transaction nodes are created. These values serve to categorise the nodes into three categories, the super nodes, random nodes and validator nodes. The super nodes are the ones that have lower latency and more computational capability. The random nodes guarantee the fairness of the network. The CNN used to classify each node is based on the AlexNet \cite{NIPS2012_4824}. This CNN receives information about the state of each node, which includes the computing power, online time, payoff and latency, and with this information it outputs the node classification. The flow of the process goes by receiving a new block data and calculating the super nodes and random nodes that create a block to be coupled to the blockchain and send that block to the nodes pool to be verified. The authors conducted many experiences with this method that show that it can be compared with PoS and PoW in terms of security, latency, costs and energy saving and still has reasonable results.

Tshilidzi Marwala and Bo Xing conducted a study about blockchain and AI \cite{Marwala2018} and present some of the problems associated with smart-contracts. The authors say that AI is the core of the new industrial revolution and that blockchain is a technology that can be integrated with AI to make it even more powerful. Based on their research, the authors infer that the blockchain is not decentralised in the sense that the underlying development is attributed to a cluster of developers and show-case the smart-contracts as an example of how this can be tragic. In the past, many hacks were done that targeted smart-contract vulnerabilities \cite{SmartHacks}. These vulnerabilities are most of the time introduced because the smart-contracts are essentially a collection of functions and data that are programmed by different human programmers. This makes the smart-contracts likely to have flaws, so, the authors show a brief overview of how AI can be used to reduce the number of bugs and flaws in the smart-contracts so that they can be delivered to the blockchain with more security. They propose an idea that can solve this problem: an AI algorithm to automatically perform formal verification of the smart-contracts to ensure a correct execution, in other words, that the smart-contracts do what they are supposed to do. The authors also introduce the concept of using computational intelligence to improve security, for example, the use of evolutionary computation to improve cryptanalysis, which can lead to the creation of more robust cyphers and consequently improving blockchain system's resilience. The authors also highlight the possible solution to the inefficiency of AI for consensus and suggest that specialised computer components to run NNs could be the solution to improve both the speed and the efficiency of this process.

\subsection{Other Proposals with Potential Impact on Robotics and AI}
\label{subsec:others}

Deeraj Nagothu et al. \cite{Nagothu2018} propose a novel approach to tackle the problem of security in traditional surveillance systems. As Smart Cities develop in today's world, the Internet-of-Things (IoT) technology becomes more important, however, it also compromises the security of the data. Traditional surveillance systems normally work with huge amounts of data and the tasks associated are growing. These tasks include people identification, aggression detection and others. Normally, these systems are designed with a centralised architecture with many servers to process the data but these designs are vulnerable to single point of failure and usually, data leaks because of the lack of protection associated with the surveillance feed. The solution proposed by the authors to tackle these problems is a system based on microservices architecture and blockchain technology. The microservices serves to isolate the video feed into different sectors, improving the system availability and robustness because the operations become decentralised. The blockchain serves as a base to synchronise the video analysis information to the microservices, proving security in a trustless network. They introduce smart contracts in their system to increase security, in order to prevent any unauthorised user from accessing the data they do not own. This offers a scalable and decentralised access control solution to the problem. By having the system distributed by microservices has the advantage of allowing the continuous development and continuous delivery without interrupting the whole service.


L.M Goodman presented in 2014 his positional paper \cite{Goodman2014} where he proposed Tezos, and later presented a white paper \cite{Goodman2016} that explained in detail all the advantages that his proposal could achieve. Tezos' base goal is to address four problems in Bitcoin, detected by the author. Those problems are the inability for Bitcoin to dynamically innovate (the "hard fork" problem), the costs and centralisation issues because of the PoW, the limited expressiveness of Bitcoin transaction language that limits smart-contracts and the security concerns regarding the implementation of a crypto-currency. So, Tezos is a platform for smart-contracts and decentralised applications that aims to have a on-chain governance in which the stakeholders can govern the protocol and decide future changes. Tezos also has improved security because it is designed to facilitate formal verification, which can mitigate many flaws in the code. This is achieved by having the code written in OCaml and by presenting a new language, Michelson \cite{Michelson}, for smart-contracts so that those contracts can be formally verified. The seed consensus is Delegated-Proof-of-Stake (DPoS) based on Slasher \cite{Slasher}, chain-of-activity \cite{Wilkinson2005} and Proof-of-Burn \cite{proof-of-burn}. With this consensus a stakeholder can delegate its Teezies (Tezos token) to other stakeholders so they can "bake" (miners are called bakers on Tezos chain) with his currency. By having a modular system that combines the transaction and the consensus protocols into a "blockchain protocol", Tezos defines the blocks in the blockchain as operators that can act on the state of the chain, allowing the blockchain protocol to become introspective, which leads to the blocks acting on the protocol itself. This is a major advantage of Tezos technology because it enables the stakeholders to vote directly on protocol upgrades to the system and avoids hard-forks. A baker is selected to create a new block when one of their Teezies is selected. This act improves the security of the system because Tezos forces the bakers to deposit Teezies to validate their work and if they act honestly (don't attempt double spends or propagation of blocks from different branches), they are rewarded, otherwise they are punished. Tezos also has endorsers. This are stakeholders that are asked to witness the creation of a block and verify its validity. This whole system can be very interesting towards improvements in robotics and AI because the self-amendment allows for changes in crucial parts of the blockchain without requiring hard-forks, the formal verification can increase the level of confidence in other participant's information, the faster consensus allows for more users to register their events onto the blockchain and it allows the creation of new AI algorithms that leverage the formal verification or the changes of the blockchain protocol.

\section{Discussion}
\label{sec:disc}

Table \ref{tab:overview} contains a short overview of all the methods that were presented in Section \ref{sec:overview}. What we take away from these proposals is that blockchain, robotics and AI are certainly going to disrupt the way we live, since they can bring so much value by themselves, but by joining them together we can compound those benefits. Blockchain has the power to store huge amounts of data that can be used to train better AI services to be used on robotics, but, the blockchain can also serve as a mechanism for transmitting information between different robots and have action-triggers coded in smart-contracts, improving the efficiency of the robots and their inter-connectivity. Although this will certainly be a fact in the near future, current methods are still in their infancy, mainly because we are going through the explosive growth phase of these technologies, and they are yet to mature.

From the methods and platforms studied, we believe that the ones that hold the most promising future are those that integrate many services in a single platform and, at the same time, share the code with the open-source community and have reward programs for finding bugs.

Certainly, we will see many robotic systems leveraging the blockchain technology, mainly in industrial and military environments where blockchain can help to automate processes with the help of smart-contracts and enable the systems to have improved security and process traceability. 
The blockchain introduces a way to trust
the data, trust other participants and to conduct internal and external changes by having certified information regarding the whole system. Scenarios where the integration of both technologies are working together to reach a common
purpose are easy to imagine. For example, a swarm of "Cop Robots" that patrol the streets greeting people and looking for miss-behaviors. These robots could communicate over the blockchain and have action-triggers with smart-contracts. These could run when they spot a person hurting another, to have the system vote on the best strategy to approach the scene or to call for help. But to achieve this type of behavior, it is necessary that smart-contracts have improved security and are able to interact with information from outside the blockchain. 

It's vital to have platforms that can integrate complementary approaches so that the market reduces from many different separated approaches to a small number of established solutions, or else, define clear interconnection standards to enable multiple solutions to talk to each other.

The marketplaces that are showing up will be crucial to make individual robots able to execute multiple complex tasks without the need for their developers to code all the different necessary solutions. This can and should be integrated with cloud robotics.

\begin{table}[]
\begin{tiny}
\caption{Short overview of the proposals discussed in the paper. Acronyms used in the table: Blockchain (BC); Smart-Contract (SC); Artificial Intelligence (AI); Proof-of-Work (PoW); Proof-of-Stake (PoS); Delegated Proof-of-Stake (DPoS); Proof-of-Concept (PoC); Proof-of-Importance (PoI); Byzantine-fault Tolerant (BFT).\label{tab:overview}}
\begin{tabular}{l|l|l|l|l|l}
\hline
\multicolumn{1}{c|}{\textbf{Name}} & \multicolumn{1}{c|}{\textbf{\begin{tabular}[c]{@{}c@{}}Problem that it's\\ solving\end{tabular}}} & \multicolumn{1}{c|}{\textbf{Solution}} & \multicolumn{1}{c|}{\textbf{Consensus}} & \multicolumn{1}{c|}{\textbf{Token}} & \multicolumn{1}{c}{\textbf{\begin{tabular}[c]{@{}c@{}}Token\\ compatibility\end{tabular}}} \\ \hline
B. Degardin \cite{BrunoDegardinLA} & \begin{tabular}[c]{@{}l@{}}Robotic Event\\ Recognition.\end{tabular} & \begin{tabular}[c]{@{}l@{}}Proprietary BC for faster\\block validation.\end{tabular} & PoW & \multicolumn{1}{c|}{-} & \multicolumn{1}{c}{-} \\ \hline
\begin{tabular}[c]{@{}l@{}}M. Fernandes and\\L. A. Alexandre\\\cite{RobotChain}\end{tabular} & \begin{tabular}[c]{@{}l@{}}Integration of \\robotics and BC,\\ mainly transaction speed\\and lack of ways to\\control such a system\\over a BC.\end{tabular} & \begin{tabular}[c]{@{}l@{}}Usage of Tezos technology\\for higher security\\and AI and SCs to\\improve performance and\\quality of robotic systems.\end{tabular} & \multicolumn{1}{c|}{-} & \multicolumn{1}{c|}{-} & \multicolumn{1}{c}{-} \\ \hline
Aitheon \cite{Aitheon} & \begin{tabular}[c]{@{}l@{}}Time-consuming\\tasks that\\developers endure.\end{tabular} & \begin{tabular}[c]{@{}l@{}}Automation with\\AI and Robotics.\end{tabular} & Multi-blind & \multicolumn{1}{c|}{AIC} & ERC-20 \\ \hline
E. C. Ferrer \cite{Ferrer2016} & \begin{tabular}[c]{@{}l@{}}Problems associated\\ with the integration of\\ BC in Swarm\\ Robotics.\end{tabular} & \begin{tabular}[c]{@{}l@{}}A set of ideas to solve\\ security issues and to\\ improve Robots performance\\ by having more information.\end{tabular} & \multicolumn{1}{c|}{-} & \multicolumn{1}{c|}{-} & \multicolumn{1}{c}{-} \\ \hline
\begin{tabular}[c]{@{}l@{}}SingularityNET\\\cite{SingularityNET} \end{tabular} & \begin{tabular}[c]{@{}l@{}}Integration of different\\ AI services so they can\\ work together seamlessly\end{tabular} & \begin{tabular}[c]{@{}l@{}}Marketplace to developers\\ sell their algorithms. An\\API that automatically\\call algorithms to solve\\defined problems.\end{tabular} & \begin{tabular}[c]{@{}l@{}}PoW\\(binded to\\Ethereum\\initially)\end{tabular} & \multicolumn{1}{c|}{AGI} & ERC-20 \\ \hline
Namahe \cite{Namahe} & \begin{tabular}[c]{@{}l@{}}The separation of the\\ different stakeholders of\\ the fashion world (sellers,\\buyers, retailers and\\ producers)\end{tabular} & \begin{tabular}[c]{@{}l@{}}A unified marketplace\\that uses AI services to\\make it more transparent\\and that automatically\\identify problems.\end{tabular} & \begin{tabular}[c]{@{}l@{}}BFT\\(based on\\IBM\\Hyperledger)\end{tabular} & \multicolumn{1}{c|}{NMH} & ERC-20 \\ \hline
Numerai \cite{Numerai} & \begin{tabular}[c]{@{}l@{}}The confidentiality\\associated with data\\from the stock market.\end{tabular} & \begin{tabular}[c]{@{}l@{}}Platform where anyone\\can download encrypted\\data about the stock market\\and be part of competitions\\with AI algorithms that in\\the end serve to guide the\\Numerai Hedge-fund.\end{tabular} & \begin{tabular}[c]{@{}l@{}}PoW\\ (based on\\Ethereum)\end{tabular} & \multicolumn{1}{c|}{NMR} & ERC-20 \\ \hline
\begin{tabular}[c]{@{}l@{}} DeepBrain Chain \\ \cite{DeepBrainChain} \end{tabular} & \begin{tabular}[c]{@{}l@{}}The lack of resources\\ to build AI services.\end{tabular} & \begin{tabular}[c]{@{}l@{}}Miners get paid to lend\\their graphics cards to train\\ neural networks instead of\\the traditional mining.\end{tabular} & \begin{tabular}[c]{@{}l@{}}DPoS and\\ PoI\end{tabular} & \multicolumn{1}{c|}{DBC} & ERC-20 \\ \hline
Neuromation \cite{Neuromation} & \begin{tabular}[c]{@{}l@{}}The lack of resources\\ to build AI services\\and the dispersion of\\models, datasets and\\AI services.\end{tabular} & \begin{tabular}[c]{@{}l@{}}A platform to sell datasets,\\models and AI services.\\ Miners get paid to lend\\their graphics cards.\end{tabular} & \begin{tabular}[c]{@{}l@{}}PoW\\(based on\\Ethereum)\end{tabular} & \multicolumn{1}{c|}{NTK} & ERC-20 \\ \hline
Eligma \cite{Eligma} & \begin{tabular}[c]{@{}l@{}}The difficulty of people\\to sell,  purchase and\\resell products.\end{tabular} & \begin{tabular}[c]{@{}l@{}}A marketplace based on\\AI services with\\product-matching and\\chatbots to help users.\end{tabular} & \begin{tabular}[c]{@{}l@{}}PoW\\(based on\\Ethereum)\end{tabular} & \multicolumn{1}{c|}{ELI} & ERC-20 \\ \hline
TraDove \cite{TraDove} & \begin{tabular}[c]{@{}l@{}}The difficulty of\\business-to-business\\contact.\end{tabular} & \begin{tabular}[c]{@{}l@{}}Marketplace that introduces\\businesses to each others\\by using AI.\end{tabular} & \multicolumn{1}{c|}{-} & \multicolumn{1}{c|}{BBC} & \multicolumn{1}{c}{-} \\ \hline
AdHive \cite{AdHive} & \begin{tabular}[c]{@{}l@{}}High expenses on the \\ marketing market.\end{tabular} & \begin{tabular}[c]{@{}l@{}}Advertisers insert\\advertisements as SCs and\\AI services matches\\them with influencers.\end{tabular} & PoC & \multicolumn{1}{c|}{ADH} & \multicolumn{1}{c}{-} \\ \hline
Matrix \cite{Matrix} & \begin{tabular}[c]{@{}l@{}}Specific languages for\\ SCs, lack of security,\\slow transaction speed\\and the inflexibility\\of the BC. \end{tabular} & \begin{tabular}[c]{@{}l@{}}A NN that automatically\\ converts simple scripts to\\ SCs, by having rules\\associated with the\\contracts, by using AI and\\by delegating the PoW.\end{tabular} & \begin{tabular}[c]{@{}l@{}}PoW and\\ PoS hybrid\end{tabular} & \multicolumn{1}{c|}{MAN} & \multicolumn{1}{c}{-} \\ \hline
ATN \cite{ATN} & \begin{tabular}[c]{@{}l@{}}Security concerns about\\ selling AI services.\end{tabular} & \begin{tabular}[c]{@{}l@{}}Marketplace based in SCs\\
to provide datasets and AI\\algorithms in a secure and\\trustworthy way.\end{tabular} & \multicolumn{1}{c|}{-} & \multicolumn{1}{c|}{ATN} & \begin{tabular}[c]{@{}l@{}}ERC-20,\\ ERC-223,\\ Qtum, \\ RSK\end{tabular} \\ \hline
Cortex \cite{Cortex} & \begin{tabular}[c]{@{}l@{}}The low integration of\\ services in SCs.\end{tabular} & \begin{tabular}[c]{@{}l@{}}An AI system that allows\\users to augment their SCs.\end{tabular} & \begin{tabular}[c]{@{}l@{}} Cuckoo\\Cycle PoW \end{tabular}& \multicolumn{1}{c|}{CTXC} & ERC-20 \\ \hline
NAM \cite{NAM} & Low-quality healthcare. & \begin{tabular}[c]{@{}l@{}}Store all medical records in\\a BC and a platform with\\AI services to give suggestions\\ and to help users access\\their data.\end{tabular} & PoS & \multicolumn{1}{c|}{NAM} & ERC-20 \\ \hline
J. Chen et al. \cite{Chen} & \begin{tabular}[c]{@{}l@{}}Problems associated\\with consensus on the\\ BC.\end{tabular} & \begin{tabular}[c]{@{}l@{}}A CNN to classify nodes\\to speed the transactions and\\lower the energy consumption\end{tabular} & AI based & \multicolumn{1}{c|}{-} & \multicolumn{1}{c}{-} \\ \hline
\begin{tabular}[c]{@{}l@{}}T. Marwala and\\B. Xing \cite{Marwala2018} \end{tabular} & \begin{tabular}[c]{@{}l@{}}Conducted a research\\about BC and AI.\end{tabular} & \begin{tabular}[c]{@{}l@{}}AI services that provide\\automatic formal verification\\of SCs and use of\\computational intelligence to\\improve security.\end{tabular} & \multicolumn{1}{c|}{-} & \multicolumn{1}{c|}{-} & \multicolumn{1}{c}{-} \\ \hline
\begin{tabular}[c]{@{}l@{}}D. Nagothu\\et al. \cite{Nagothu2018} \end{tabular}& \begin{tabular}[c]{@{}l@{}}Security in traditional\\ surveillance systems\end{tabular} & \begin{tabular}[c]{@{}l@{}}System based on microservices\\ architecture and BC technology\end{tabular} & \multicolumn{1}{c|}{-} & \multicolumn{1}{c|}{-} & \multicolumn{1}{c}{-} \\ \hline
\begin{tabular}[c]{@{}l@{}}L.M Goodman\\ \cite{Goodman2014,Goodman2016}\end{tabular} & \begin{tabular}[c]{@{}l@{}}Inability of BC to \\ dynamically innovate;\\ Costs and \\centralisation issues;\\ Limited expressiveness \\ of SCs languages;\\ Security concerns about\\ tokens\end{tabular} & \begin{tabular}[c]{@{}l@{}}Tezos, a self-amending,\\permissionless,\\ distributed platform;\\Stakeholders decide the\\future of the platform;\\Modular BC to avoid\\hard-forks; \\ Formal verification.\end{tabular} & \multicolumn{1}{c|}{-} & \multicolumn{1}{c|}{XTZ} & \begin{tabular}[c]{@{}l@{}}Ether;\\ Bitcoin;\\Teezies\end{tabular} \\ \hline
\end{tabular}
\end{tiny}
\end{table}

\section{Conclusions}
\label{sec:conc}

Blockchain technology is still in its infancy and its possible impact on the global
economy is yet to be clearly understood. The integration of services with the
blockchain, specially robotics, is still in an early prototype stage. This means
that many improvements are conducted in separated blockchains. There aren't
yet clear winner technologies and most market participants are not aware
of the new technologies and sometimes lack confidence on the robustness of these
first proposals. 
Proposed approaches are abundant, interconnection standards are missing and the integration of those approaches with Industry 4.0 or cloud robotics, e.g., is yet to be achieved.

In this paper, we overviewed many of the current methods and proposals for
the blockchain technology that either use robotics or leverage AI services that
can be used to improve robotic systems. We also studied some proposals that try
to enhance the blockchain technologies and that can have an important impact
on systems like swarm robotics, by giving them increased security and trust on
the blockchain data.

As this emerging technology, that will have a profound impact in society, is maturing it will interact with many other paradigms, such as robotics and AI, to yield improved products and productivity, services and higher living standards for our society.

\bibliographystyle{splncs04}
\bibliography{bibliografia}

\end{document}